\documentclass{article}
\usepackage{ismir,amsmath,cite,url}
\usepackage{graphicx}
\usepackage{color}
\usepackage{amsfonts,amssymb}
\usepackage{booktabs}

\newcommand{\E}{\mathbb{E}}
\newcommand{\R}{\mathbb{R}}
\newcommand{\cR}{\mathcal{R}}
\newcommand{\etal}{\textit{et al.}}

\title{automatic, personalized, and flexible playlist generation using reinforcement learning}

\twoauthors
{Shun-Yao Shih} {National Taiwan University \\ shunyaoshih@gmail.com}
{Heng-Yu Chi} {KKBOX Inc., Taipei, Taiwan \\ henrychi@kkbox.com}

\begin{document}

\maketitle
\begin{abstract}
  Songs can be well arranged by professional music curators to form a riveting
  playlist that creates engaging listening experiences. However, it is
  time-consuming for curators to timely rearrange these playlists for
  fitting trends in future.
  By exploiting the techniques of deep learning and reinforcement learning,
  in this paper, we consider music playlist generation as a language modeling problem
  and solve it by the proposed attention language model with policy gradient.
  We develop a systematic and interactive approach
  so that the resulting playlists can be tuned flexibly according to user preferences.
  Considering a playlist as a sequence of words,
  we first train our attention RNN language model on
  baseline recommended playlists.
  By optimizing suitable imposed reward functions,
  the model is thus refined for corresponding preferences.
  The experimental results demonstrate that our approach
  not only generates coherent playlists automatically
  but is also able to flexibly recommend personalized playlists
  for diversity, novelty and freshness.
\end{abstract}

\section{Introduction}\label{sec:introduction}
Professional music curators or DJs are usually able to carefully select, order,
and form a list of songs which can give listeners brilliant listening experiences.
For a music radio with a specific topic, they can collect songs related to the topic
and sort in a smooth context.
By considering preferences of users,
curators can also find what they like and recommend them several lists of songs.
However, different people have different preferences toward diversity, popularity, and etc.
Therefore,
it will be great if we can refine playlists based on different preferences of users on the fly.
Besides, as online music streaming services grow,
there are more and more demands for efficient and effective music playlist recommendation.
Automatic and personalized music playlist generation thus becomes a critical issue.

However, it is unfeasible and expensive for editors
to daily or hourly generate suitable playlists for all users
based on their preferences about trends, novelty, diversity, etc.
Therefore, most of previous works try to deal with such problems
by considering some particular assumptions.
McFee $\etal$\cite{2011} consider playlist generation as
a language modeling problem and solve it by adopting statistical techniques.
Unfortunately,
statistical method does not perform well on small datasets.
Pampalk $\etal$\cite{2005} generate playlists by exploiting explicit user behaviors
such as skipping.
However, for implicit user preferences on playlists,
they do not provide a systematic way to handle it.

As a result, for generating personalized playlists automatically and flexibly,
we develop a novel and scalable music playlist generation system.
The system consists of three main steps.
First, we adopt Chen $\etal$'s work~\cite{songnn}
to generate baseline playlists based on the preferences of users about songs.
In details,
given the relationship between users and songs,
we construct a corresponding bipartite graph at first.
With the users and songs graph,
we can calculate embedding features of songs and thus
obtain the baseline playlist for each songs by
finding their k-nearest neighbors.
Second, by formulating baseline playlists as sequences of words,
we can pretrain RNN language model (RNN-LM) to obtain better initial parameters
for the following optimization, using policy gradient reinforcement learning. 
We adopt RNN-LM\@ because not only
RNN-LM\@ has better ability of learning information progresses
than traditional statistical methods in many generation tasks,
but also neural networks can be combined with reinforcement learning
to achieve better performances\cite{rl_chatbot}.
Finally, given preferences from user profiles and the pretrained parameters,
we can generate personalized playlists by exploiting techniques of
policy gradient reinforcement learning
with corresponding reward functions.
Combining these training steps,
the experimental results show that
we can generate personalized playlists to satisfy different preferences of users with ease.

Our contributions are summarized as follows:
\begin{itemize}
  \item We design an automatic playlist generation framework,
    which is able to provide timely recommended playlists for online music streaming services.
  \item We remodel music playlist generation into a sequence prediction problem using RNN-LM\@
    which is easily combined with policy gradient reinforcement learning method.
  \item The proposed method can flexibly generate suitable personalized playlists
    according to user profiles using corresponding optimization goals in policy gradient.
\end{itemize}

The rest of this paper is organized as follows.
In Section~\ref{sec:related_work},
we introduce several related works about playlist generation and recommendation.
In Section~\ref{sec:policy_gradient},
we provide essential prior knowledge of our work related to policy gradient.
In Section~\ref{sec:model_introduction},
we introduce the details of our proposed model,
attention RNN-LM\@ with concatenation (AC-RNN-LM).
In Section~\ref{sec:experiments},
we show the effectiveness of our method and
conclude our work in Section~\ref{sec:conclusions}.

\section{Related Work}\label{sec:related_work}

Given a list of songs, previous works try to rearrange them for better song sequences~\cite{2001, 2002, 2008, 2017}.
First, they construct a song graph by considering songs in playlist as vertices, and relevance of audio features between songs as edges.
Then they find a Hamiltonian path with some properties, such as smooth transitions of songs~\cite{2017}, to create new sequencing of songs.
User feedback is also an important consideration when we want to generate playlists~\cite{2005, 2009, topics, hits}.
%%%%%
By considering several properties, such as tempo, loudness, topics, and artists, 
of users' favorite played songs recently,
authors of~\cite{topics, hits} can thus provide personalized playlist for users based on favorite properties of users.
%%%%%
Pampalk $\etal$~\cite{2005} consider skip behaviors as negative signals and
the proposed approach can automatically choose the next song according to
audio features and avoid skipped songs at the same time.
Maillet $\etal$~\cite{2009} provides a more interactive way to users.
Users can manipulate weights of tags to express high-level music characteristics and
obtain corresponding playlists they want.
% new {{{
% Instead of explicitly considering user feedback,
% some works tried to automatically provide new playlists for each user
% based on the feature analysis of user's recently played songs
% such as tempo, loudness, topics, and artists\cite{topics, hits}.
% }}}
To better integrate user behavior into playlist generation,
several works are proposed to combine playlist generation algorithms
with the techniques of reinforcement learning\cite{CFRL, DJMC}.
Xing $\etal$ first introduce exploration into traditional collaborative filtering
to learn preferences of users.
Liebman $\etal$ take the formulation of
Markov Decision Process into playlist generation framework
to design algorithms that learn representations for preferences of users
based on hand-crafted features.
By using these representations, they can generate personalized playlist for users.

Beyond playlist generation,
there are several works adopting the concept of playlist generation to facilitate recommendation systems.
Given a set of songs, Vargas $\etal$~\cite{algorithm1} propose several scoring functions,
such as diversity and novelty, and
retrieve the top-K songs with higher scores for each user as the resulting recommended list of songs.
Chen $\etal$~\cite{songnn} propose a query-based music recommendation system
that allow users to select a preferred song as a seed song to obtain related songs as a recommended playlist.

\section{Policy Gradient Reinforcement Learning}\label{sec:policy_gradient}

Reinforcement learning has got a lot of attentions from public
since Silver $\etal$\cite{alphazero} proposed a general reinforcement learning algorithm
that could make an agent achieve superhuman performance in many games.
Besides, reinforcement learning has been successfully applied to many other problems
such as dialogue generation modeled as Markov Decision Process (MDP).

A Markov Decision Process is usually denoted by a tuple $\mathcal{(S, A, P, R, \gamma)}$, where
\begin{itemize}
  \item $\mathcal{S}$ is a set of states
\vspace{-5pt}
  \item $\mathcal{A}$ is a set of actions
\vspace{-5pt}
  \item $\mathcal{P}(s, a, s') = \Pr[s' | s, a]$ is
    the transition probability that action $a$ in state $s$ will lead to state $s'$
\vspace{-5pt}
  \item $\cR(s, a) = \E[r | s, a]$ is
    the expected reward that an agent will receive when the agent does action $a$ in state $s$.
\vspace{-5pt}
  \item $\mathcal{\gamma} \in [0, 1]$ is
    the discount factor representing the importance of future rewards
\end{itemize}

Policy gradient is a reinforcement learning algorithm to solve MDP\@ problems.
Modeling an agent with parameters $\theta$,
the goal of this algorithm is to find the best $\theta$ of
a policy $\pi_{\theta}(s, a) = \Pr[a | s, \theta]$
measured by average reward per time-step

\vspace{-5pt}
\begin{equation}\label{policy_gradient_goal}
  J(\theta) = \sum_{s \in \mathcal{S}} d^{\pi_{\theta}}(s) \sum_{a \in \mathcal{A}} \pi_{\theta}(s, a) \cR(s, a)
\end{equation}
where $d^{\pi_{\theta}}(s)$ is stationary distribution of Markov chain for $\pi_{\theta}$.

Usually, we assume that $\pi_{\theta}(s, a)$ is differentiable with respect to
its parameters $\theta$, i.e., $\frac{\partial \pi_{\theta}(s, a)}{\partial \theta}$ exists,
and solve this optimization problem \eqnref{policy_gradient_goal} by gradient ascent.
Formally, given a small enough $\alpha$, we update its parameters $\theta$ by

\vspace{-10pt}
\begin{equation}
  \theta \leftarrow \theta + \alpha \nabla_{\theta} J(\theta)
\end{equation}
\vspace{-5pt}
where
\begin{equation}
  \begin{split}
    \nabla_{\theta} J(\theta) &=\sum_{s \in \mathcal{S}} d^{\pi_{\theta}}(s) \sum_{a \in \mathcal{A}} \pi_{\theta}(s, a) \nabla_{\theta} \pi_{\theta}(s, a) \cR(s, a) \\
                              &= \mathbb{E}[\nabla_{\theta} \pi_{\theta}(s, a) \cR(s, a)]
  \end{split}
\end{equation}
\vspace{-10pt}

\begin{figure*}[ht]
 \includegraphics[width=2.0\columnwidth]{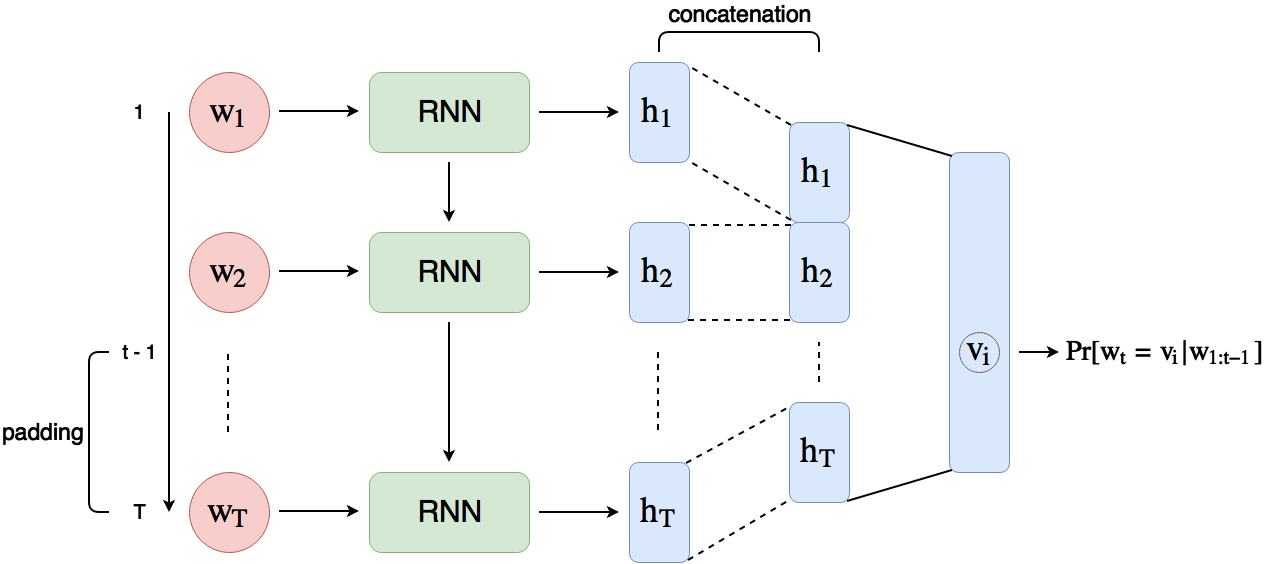}
 \caption{The structure of our attention RNN\@ language model with concatenation}
\label{fig:model}
\end{figure*}

\vspace{-10pt}

\section{The Proposed Model}\label{sec:model_introduction}
The proposed model consists of two main components.
We first introduce the structure of the proposed RNN-based model
in Section~\ref{ssec:lm}.
Then in Section~\ref{ssec:rl},
we formulate the problem as a Markov Decison Process
and solve the formulated problem by policy gradient to generate refined playlists.

\subsection{Attention RNN Language Model}\label{ssec:lm}
Given a sequence of tokens $\{w_1, w_2, \ldots, w_t\}$,
an RNN-LM\@ estimates the probability $\Pr[w_t | w_{1:t - 1}]$ with a recurrent function

\vspace{-5pt}
\begin{equation}
  h_t = f(h_{t - 1}, w_{t - 1})
\end{equation}
\vspace{-5pt}
\\
and an output function, usually softmax,

\vspace{-5pt}
\begin{equation}\label{output_function}
  \Pr[w_t = v_i | w_{1:t - 1}] =
  \frac{\exp({W_{v_i}^{\top} h_t + b_{v_i}})}{\sum_k \exp({W_{v_k}^\top h_t + b_{v_k}})}
\end{equation}
\vspace{-5pt}
\\
where the implementation of the function $f$ depends on which kind of RNN\@ cell we use,
$h_t \in \R^D$, $W \in \R^{D \times V}$
with the column vector $W_{v_i}$ corresponding to a word $v_i$,
and $b \in \R^V$ with the scalar $b_{v_i}$ corresponding to a word $v_i$
($D$ is the number of units in RNN\@, and $V$ is the number of unique tokens in all sequences).

We then update the parameters of the RNN-LM
by maximizing the log-likelihood on a set of sequences with size $N$, $\{s_1, s_2, \ldots, s_N\}$,
and the corresponding tokens, $\{w_1^{s_i}, w_2^{s_i}, \ldots, w_{|s_i|}^{s_i}\}$.

\vspace{-5pt}
\begin{equation}
  \mathcal{L} = \frac{1}{N} \sum_{n = 1}^N \sum_{t = 2}^{|s_n|} \log\Pr[w_t^{s_n} | w_{1: t - 1}^{s_n}]
\end{equation}

\subsubsection{Attention in RNN-LM}
Attention mechanism in sequence-to-sequence model has been proven to be effective in the fields of
image caption generation, machine translation, dialogue generation, and etc.
Several previous works also indicate that attention is even more impressive
on RNN-LM\cite{ARNNLM}.

In attention RNN language model (A-RNN-LM), given the hidden states
from time $t - C_{ws}$ to $t$, denoted as $h_{t - C_{ws}: t}$,
where $C_{ws}$ is the attention window size,
we want to compute a context vector $c_t$ as
a weighted sum of hidden states $h_{t - C_{ws}: t - 1}$ and then
encode the context vector $c_t$ into the original hidden state $h_t$.

\vspace{-5pt}
\begin{equation}
  \beta_i = \nu^\top \tanh(W_1 h_t + W_2 h_{t - C_{ws} + i})
\end{equation}
\vspace{-5pt}
\begin{equation}
  \alpha_i = \frac{\exp(\beta_i)}{\sum_{k=0}^{C_{ws} - 1} \exp(\beta_k)}
\end{equation}
\vspace{-5pt}
\begin{equation}
  c_t = \sum_{i=0}^{C_{ws} - 1} \alpha_i h_{t - C_{ws} + i}
\end{equation}
\vspace{-5pt}
\begin{equation}
  h'_t = W_3
  \begin{bmatrix}
    h_t \\ c_t
  \end{bmatrix}
\end{equation}
\vspace{-5pt}
\\
where $\beta$ is Bahdanau's scoring style\cite{attention}, $W_1, W_2 \in \R^{D \times D}$,
and $W_3 \in \R^{D \times 2D}$.

\subsubsection{Our Attention RNN-LM\@ with concatenation}\label{ssec:our_work}
In our work, $\{s_1, s_2, \ldots, s_N\}$ and
$\{w_1^{s_i}, w_2^{s_i}, \ldots, w_{|s_i|}^{s_i}\}$
are playlists and songs by adopting Chen $\etal$'s work\cite{songnn}.
More specifically, given a seed song $w_1^{s_i}$ for a playlist $s_i$,
we find top-k approximate nearest neighbors of $w_1^{s_i}$
to formulate a list of songs $\{w_1^{s_i}, w_2^{s_i}, \ldots, w_{|s_i|}^{s_i}\}$.

The proposed attention RNN-LM\@ with concatenation (AC-RNN-LM) is shown in \figref{fig:model}.
We pad $w_{1: t - 1}$ to $w_{1:T}$
and concatenate the corresponding $h'_{1: T}$
as the input of our RNN-LM's output function in \eqnref{output_function},
where $T$ is the maximum number of songs we consider in one playlist.
Therefore, our output function becomes

\vspace{-10pt}
\begin{equation}
  \Pr[w_t = v_i | w_{1:t - 1}] =
  \frac{\exp(W_{v_i}^{\top} h' + b_{v_i})}{\sum_k \exp(W_{v_k}^\top h' + b_{v_k})}
\end{equation}
\vspace{-5pt}
\\
where $W \in \R^{DT \times V}$, $b \in \R^{V}$ and
\begin{equation}
  h' =
  \begin{bmatrix}
    h'_1 \\ h'_2 \\ \vdots \\ h'_T
  \end{bmatrix}
  \in \R^{DT \times 1}
\end{equation}
\vspace{-5pt}

\subsection{Policy Gradient}\label{ssec:rl}

We exploit policy gradient in order to optimize \eqnref{policy_gradient_goal}, which is formulated as follows.

\subsubsection{Action}
An action $a$ is a song id, which is a unique representation of each song, that the model is about to generate.
The set of actions in our problem is finite
since we would like to recommend limited range of songs.
\subsubsection{State}
A state $s$ is the songs we have already recommended including the seed song,
$\{w_1^{s_i}, w_2^{s_i}, \ldots, w_{t - 1}^{s_i}\}$.
\subsubsection{Policy}
A policy $\pi_{\theta}(s, a)$ takes the form of our AC-RNN-LM and
is defined by its parameters $\theta$.
\subsubsection{Reward}
Reward $\cR(s, a)$ is a weighted sum of several reward functions, i.e.,
$\cR_i: s \times a \mapsto \R$. 
In the following introductions,
we formulate 4 important metrics for playlists generation.
The policy of our pretrained AC-RNN-LM\@ is denoted as $\pi_{\theta_{RNN}}(s, a)$
with parameters $\theta_{RNN}$, and
the policy of our AC-RNN-LM\@ optimized by policy gradient is denoted
as $\pi_{\theta_{RL}}(s, a)$ with parameters $\theta_{RL}$.

\begin{description}
  \item [Diversity] represents the variety in a recommended list of songs.
    Several generated playlists in Chen $\etal$'s work\cite{songnn} are composed of songs
    with the same artist or album.
    It is not regarded as a good playlist for recommendation system because of low diversity.
    Therefore, we formulate the measurement of the diversity by the euclidean distance
    between the embeddings of the last song in the existing playlist, $w_{|s|}^s$, and
    the predicted song, $a$.
    \begin{equation}\label{R_1}
      \cR_1(s, a) = -\log(|d(w_{|s|}^s, a) - C_{\text{distance}}|)
    \end{equation}
    \\
    where $d(\cdot)$ is the euclidean distance between the embeddings of $w_{|s|}^s$ and $a$,
    and $C_{\text{distance}}$ is a parameter that represents
    the euclidean distance that we want the model to learn.
    
  \item [Novelty] is also important for a playlist generation system.
    We would like to recommend something new to users instead of recommend something familiar.
    Unlike previous works, our model can easily generate playlists with novelty
    by applying a corresponding reward function.
    As a result, we model reward of novelty as
    a weighted sum of normalized playcounts in periods of time\cite{auralist}.
    \begin{equation}\label{R_2}
      \cR_2(s, a) = -\log(\sum_t w(t) \frac{\log(p_t(a))}{\log(\max_{a' \in A} p_t(a'))})
    \end{equation}
    \\
    where $w(t)$ is the weight of a time period, $t$, with a constraint $\sum_t w(t) = 1$,
    $p_t(a)$ is playcounts of the song $a$, and $A$ is the set of actions.
    Note that songs with less playcounts have higher value of $\cR_2$, and vice versa.
    
  \item [Freshness] % 看看有沒有更好的字
    is a subjective metric for personalized playlist generation.
    For example, latest songs is usually more interesting for young people,
    while older people would prefer old-school songs.
    Here, we arbitrarily choose one direction for optimization to the agent $\pi_{\theta_{RL}}$
    to show the feasibility of our approach.
    \begin{equation}\label{R_3}
      \cR_3(s, a) = -\log(\frac{Y_{a} - 1900}{2017 - 1900})
    \end{equation}
    \\
    where $Y_a$ is the release year of the song $a$.
    
  \item [Coherence] % coherent playlists?
    is the major feature we should consider to avoid situations that
    the generated playlists are highly rewarded but lack of cohesive listening experiences.
    We therefore consider the policy of our pretrained language model,
    $\pi_{\theta_{RNN}}(s, a)$, which is well-trained on coherent playlists,
    as a good enough generator of coherent playlists.

    \begin{equation}\label{R_4}
      \cR_4(s, a) = \log(\Pr[a | s, \theta_{RNN}])
    \end{equation}
\end{description}

Combining the above reward functions, our final reward for the action $a$ is
\begin{equation}
  \begin{split}
    \cR(s, a) = &\gamma_1 \cR_1(s, a) + \gamma_2 \cR_2(s, a) + \\
                &\gamma_3 \cR_3(s, a) + \gamma_4 \cR_4(s, a)
  \end{split}
  \label{rewards}
\end{equation}
\\
where the selection of $\gamma_1$, $\gamma_2$, $\gamma_3$, and $\gamma_4$
depends on different applications.

Note that
although we only list four reward functions here,
the optimization goal $\cR$ can be easily extended
by a linear combination of more reward functions.

\section{Experiments and Analysis}\label{sec:experiments}

In the following experiments, we first introduce the details of dataset and evaluation metrics in Section~\ref{ssec:metrics} and training details in Section~\ref{ssec:training_details}.
In Section~\ref{ssec:pretrain}, we compare pretrained RNN-LMs\@ with different mechanism combination
by perplexity to show our proposed AC-RNN-LM\@ is
more effectively and efficiently than others.
In order to demonstrate the effectiveness of our proposed method,
AC-RNN-LM\@ combined with reinforcement learning,
we adopt three standard metrics, diversity, novelty, and freshness
(cf. Section ~\ref{ssec:metrics}) to validate our models in Section~\ref{sssec:rl_results}.
Moreover, we demonstrate that it is effortless to flexibly manipulate
the properties of resulting generated playlists in Section~\ref{sssec:gamma}.
Finally, in Section~\ref{sssec:limitation},
we discuss the details about the design of reward functions with given preferred properties.

\begin{figure}
 \includegraphics[width=\columnwidth]{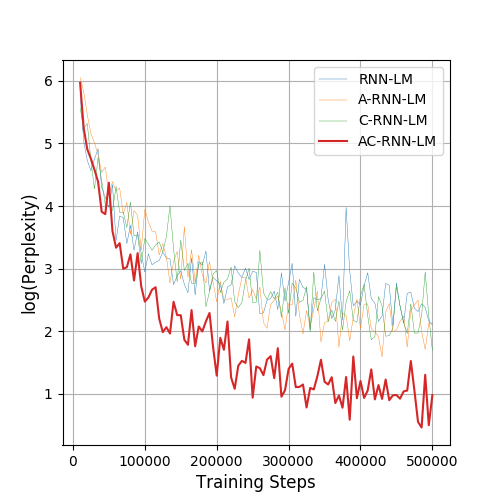}
 \caption{Log-perplexity of different pretrained models on the dataset
   under different training steps
 }\label{fig:model_comparison}
\end{figure}

\subsection{Dataset and Evaluation Metrics}\label{ssec:metrics}
The playlist dataset is provided by KKBOX Inc.,
which is a regional leading music streaming company.
It consists of $10,000$ playlists,
each of which is composed of $30$ songs.
There are $45,243$ unique songs in total.

For validate our proposed approach, we use the metrics as follows.
\vspace{-1pt}
\begin{description}
  \item [Perplexity] is calculated based on the song probability distributions, which is shown as follows.
    \begin{equation*}
      perplexity = e^{\frac{1}{N} \sum_{n=1}^N \sum_x -q(x) \log p(x)}
    \end{equation*}
    where $N$ is the number of training samples, $x$ is a song in our song pool,
    $p$ is the predicted song probability distribution, and
    $q$ is the song probability distribution in ground truth.
    \vspace{-1pt}
  \item [Diversity] is computed as different unigrams of artists scaled by 
  he total length of each playlist, which is measured by Distinct-1\cite{unigram}
    \vspace{-1pt}
  \item [Novelty] is designed for recommending something new to users~\cite{auralist}.
    The more the novelty is, the lower the metric is.
  \vspace{-1pt}
  \item [Freshness] is directly measured by
    the average release year of songs in each playlist.
\end{description}

\subsection{Training Details}\label{ssec:training_details}
In the pretraining and reinforcement learning stage,
we use 4 layers and 64 units per layer in all RNN-LM\@ with LSTM units, and
we choose $T = 30$ for all RNN-LM\@ with padding and concatenation.
The optimizer we use is Adam\cite{adam}.
The learning rates for pretraining stage and reinforcement learning stage
are empirically set as 0.001 and 0.0001, respectively.

\subsection{Pretrained Model Comparison}\label{ssec:pretrain}
In this section,
we compare the training error of RNN-LM\@ combining with different mechanisms.
The RNN-LM\@ with attention is denoted as A-RNN-LM, 
the RNN-LM\@ with concatenation is denoted as C-RNN-LM, and
the RNN-LM\@ with attention and concatenation
is denoted as AC-RNN-LM\@.
\figref{fig:model_comparison} reports the training error of different RNN-LMs\@ as
log-perplexity which is equal to negative log-likelihood
under the training step from $1$ to $500,000$.
Here one training step means that we update our parameters by one batch.
As shown in \figref{fig:model_comparison},
the training error of our proposed model, AC-RNN-LM\@, can
not only decrease faster than the other models
but also reach the lowest value at the end of training.
Therefore,
we adopt AC-RNN-LM\@ as our pretrained model.

Worth noting that the pretrained model is developed for two purposes. 
One is to provide a good basis for further optimization, and
the other is to estimate transition probabilities of songs in the reward function.
Therefore, we simply select the model with the lowest training error
to be optimized by policy gradient and an estimator of $\Pr[a | s, \theta_{RNN}]$
(cf. \eqnref{R_4}).

\begin{table}[]
\centering
\caption{Weights of reward functions for each model}
\label{rl_weights}
\begin{tabular}{@{}ccccc@{}}
\toprule
Model      & $\gamma_1$ & $\gamma_2$ & $\gamma_3$ & $\gamma_4$ \\ \midrule
RL-DIST    & 0.5        & 0.0        & 0.0        & 0.5        \\
RL-NOVELTY & 0.0        & 0.5        & 0.0        & 0.5        \\
RL-YEAR    & 0.0        & 0.0        & 0.5        & 0.5        \\
RL-COMBINE & 0.2        & 0.2        & 0.2        & 0.4        \\ \bottomrule
\end{tabular}
\end{table}
\begin{table}[]
\centering
\caption{Comparison on different metrics for playlist generation system
 (The bold text represents the best, and the underline text represents the second)}
\label{rl_metrics}
\begin{tabular}{@{}cccc@{}}
\toprule
Model       & Diversity        & Novelty          & Freshness           \\ \midrule
Embedding\cite{songnn}   & 0.32             & 0.19             & 2007.97             \\
AC-RNN-LM   & 0.39             & 0.20             & 2008.41             \\
RL-DIST     & \underline{0.44} & 0.20             & 2008.37             \\
RL-NOVELTY  & 0.42             & \textbf{0.05}    & 2012.89             \\
RL-YEAR     & 0.40             & 0.19             & \textbf{2006.23}    \\
RL-COMBINE  & \textbf{0.49}    & \underline{0.18} & \underline{2007.64} \\ \bottomrule
\end{tabular}
\end{table}

\begin{figure}[t]
    \centering
    \includegraphics[width=\columnwidth]{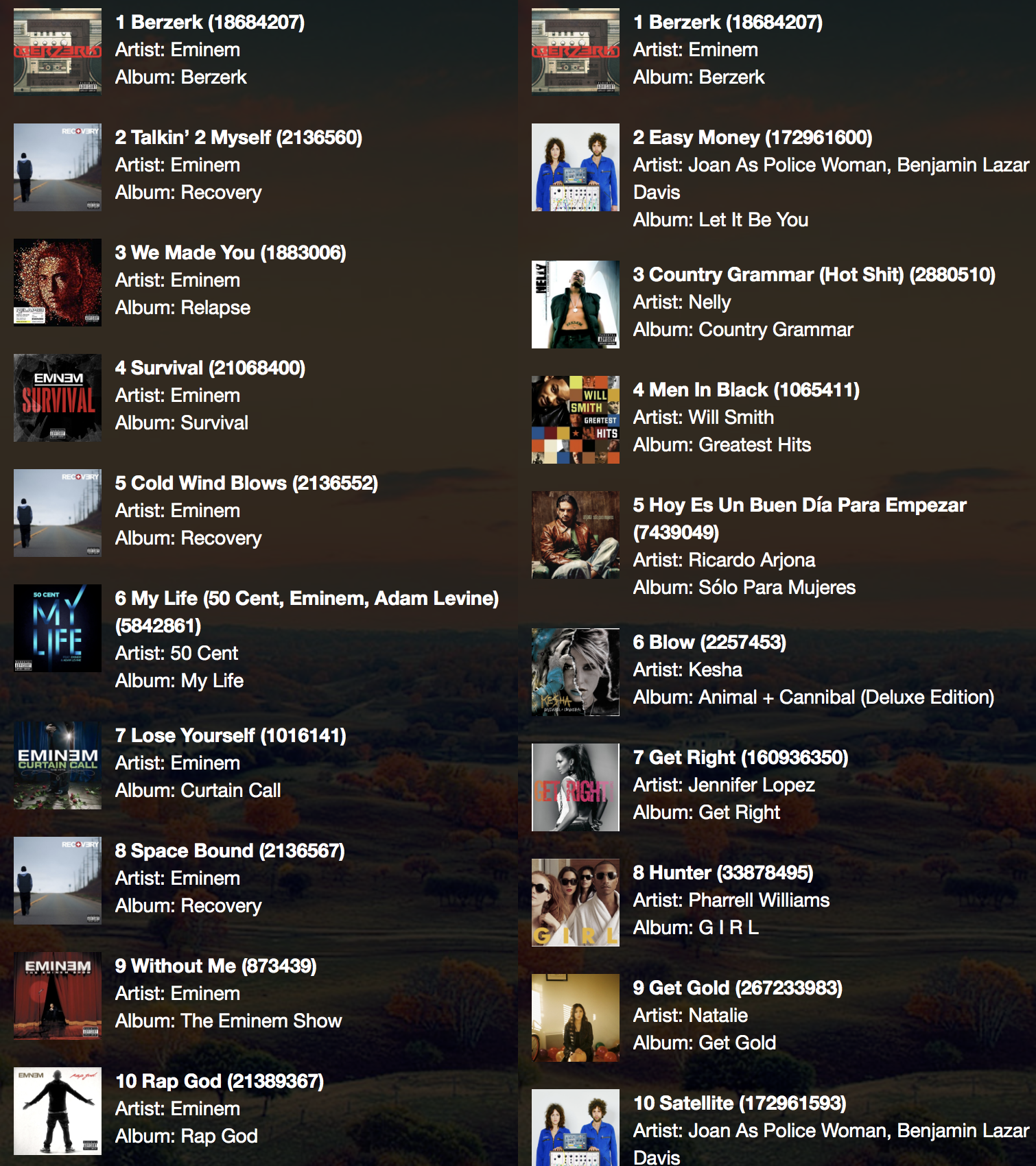}
    \caption{Sample playlists generated by our approach. The left one is generated by Embedding\cite{songnn} and
    the right one is generated by RL-COMBINE.}
    \vspace{-10pt}
    \label{fig:sample_playlist}
\end{figure}

\subsection{Playlist Generation Results}\label{sssec:rl_results}
As shown in Table~\ref{rl_metrics},
to evaluate our method,
we compare 6 models on 3 important features,
diversity, novelty, and freshness (cf. Section~\ref{ssec:metrics}),
of playlist generation system.
The details of models are described as follows.
Embedding represents the model of Chen $\etal$'s work\cite{songnn}.
Chen $\etal$ construct the song embedding by relationships between user and song and
then finds approximate $k$ nearest neighbors for each song.
RL-DIST, RL-NOVELTY, RL-YEAR, and RL-COMBINE are models
that are pretrained 
and optimized by the policy gradient method (cf. \eqnref{rewards}) with different weights, respectively, as shown in Table~\ref{rl_weights}.

The experimental results show that for single objective such as diversity, our models can accurately generate playlists with corresponding property.
For example, RL-Year can generate a playlist
which consists of songs with earliest release years
than Embedding and AC-RNN-LM\@.
Besides, even when we impose our model with multiple reward functions,
we can still obtain a better resulting playlist in comparison with Embedding and AC-RNN-LM\@.
Sample result is shown in \figref{fig:sample_playlist}.

From Table~\ref{rl_metrics},
we demonstrate that by using appropriate reward functions,
our approach can generate playlists
to fit the corresponding needs easily.
We can systematically find more songs from different artists (RL-DIST),
more songs heard by fewer people (RL-NOVELTY),
or more old songs for some particular groups of users (RL-YEAR).

\subsection{Flexible Manipulating Playlist Properties}\label{sssec:gamma}

After showing that our approach can easily fit several needs,
we further investigate the influence of $\gamma$ to the resulting playlists.
In this section,
several models are trained with the weight $\gamma_2$ from $0.0$ to $1.0$
to show the variances in novelty of the resulting playlists.
Here we keep $\gamma_2 + \gamma_4 = 1.0$ and $\gamma_1 = \gamma_3 = 0$ and
fix the training steps to $10,000$.

\begin{figure}
 \includegraphics[width=\columnwidth]{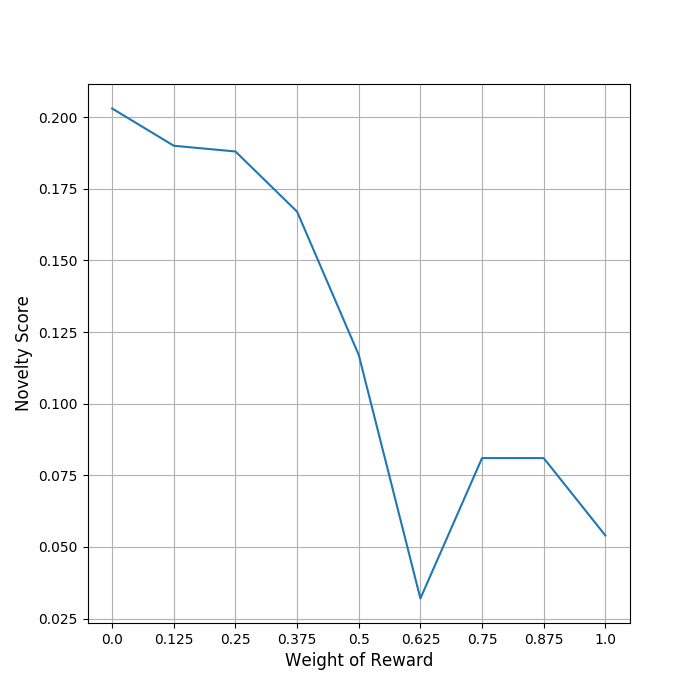}
 \caption{Novelty score of playlists generated by different imposing weights}
 \vspace{-10pt}
 \label{fig:pop_progression}
\end{figure}

As shown in \figref{fig:pop_progression},
novelty score generally decreases when $\gamma_2$ increases from $0.0$ to $1.0$
but
it is also possible that
the model may sometimes find the optimal policy earlier than expectation
such as the one with $\gamma_2 = 0.625$.
Nevertheless, in general,
our approach can not only let the model generate more novel songs
but also make the extent of novelty be controllable.
Besides automation, this kind of flexibility is also important in applications.

Take online music streaming service as an example,
when the service provider wants to recommend playlists to a user
who usually listens to non-mainstream but familiar songs (i.e., novelty score is $0.4$),
it is more suitable to generate playlists which
consists of songs with novelty scores equals to $0.4$
instead of generating playlists which is composed of $60\%$ songs with novelty scores equals to $0.0$
and $40\%$ songs with novelty scores equals to $1.0$.
Since users usually have different kinds of preferences on each property,
to automatically generate playlists fitting needs of each user,
such as novelty, becomes indispensable.
The experimental results verify that our proposed approach can satisfy the above application.

\subsection{Limitation on Reward Function Design}\label{sssec:limitation}
When we try to define a reward function $\cR_i$ for a property,
we should carefully avoid the bias from the state $s$.
In other words,
reward functions should be specific to the corresponding feature we want.
One common issue is that
the reward function may be influenced by the number of songs in state $s$.
For example, in our experiments,
we adopt Distinct-1 as the metric for diversity.
However, we cannot also adopt Distinct-1 as our reward function directly
because it is scaled by the length of playlists
which results in all actions from states with fewer songs will be benefited.
Therefore, difference between $cR_1$ and Distinct-1 is the reason that
RL-DIST does not achieve the best performance in Distinct-1 (cf. Table~\ref{rl_weights}). 
In summary,
we should be careful to design reward functions, and
sometimes we may need to formulate another approximation objective function to avoid biases.

\vspace{-10pt}
\section{Conclusions and Future Work}\label{sec:conclusions}
In this paper, we develop a playlist generation system 
which is able to generate personalized playlists automatically and flexibly.
We first formulate playlist generation as a language modeling problem.
Then by exploiting the techniques of RNN-LM\@ and reinforcement learning, 
the proposed approach can flexibly generate suitable playlists
for different preferences of users.

In our future work,
we will further investigate the possibility to
automatically generate playlists by considering qualitative feedback.
For online music streaming service providers,
professional music curators will give qualitative feedback on generated playlists
so that research developers can improve the quality of playlist generation system.
Qualitative feedback such as `songs from diverse artists but similar genres'
is easier to be quantitative.
We can design suitable reward functions and generate corresponding playlists by our approach. 
However, other feedback such as `falling in love playlist'
is more difficult to be quantitative.
Therefore, we will further adopt audio features and explicit tags of songs 
in order to provide a better playlist generation system.

% For bibtex users:
\bibliography{ISMIRtemplate}

% For non bibtex users:
%\begin{thebibliography}{citations}
%
%\bibitem {Author:00}
%E. Author.
%``The Title of the Conference Paper,''
%{\it Proceedings of the International Symposium
%on Music Information Retrieval}, pp.~000--111, 2000.
%
%\bibitem{Someone:10}
%A. Someone, B. Someone, and C. Someone.
%``The Title of the Journal Paper,''
%{\it Journal of New Music Research},
%Vol.~A, No.~B, pp.~111--222, 2010.
%
%\bibitem{Someone:04} X. Someone and Y. Someone. {\it Title of the Book},
%    Editorial Acme, Porto, 2012.
%
%\end{thebibliography}

\end{document}